\documentclass[10 pt, conference]{support/ieeeconf}
\usepackage{times}
\usepackage[pdftex]{graphicx}
\usepackage{subfigure}
\usepackage{amsmath,amssymb,amsopn,amstext,amsfonts}
\usepackage{cancel}
\usepackage[space]{cite}
\usepackage{pdfsync}
\usepackage{balance}
\usepackage{color}
\usepackage{mathtools}
\usepackage{algpseudocode}
\usepackage{algorithm} 

\algnewcommand\algorithmicforeach{\textbf{for each}}
\algdef{S}[FOR]{ForEach}[1]{\algorithmicforeach\ #1\ \algorithmicdo}

\usepackage{bm}

\usepackage{diagbox}
\usepackage{float}
\usepackage{epstopdf}
\usepackage{pifont}
\usepackage{fixltx2e}
\usepackage{mathrsfs}
\usepackage{booktabs}
\usepackage{multirow}
\usepackage{tabularx}
\usepackage{url}
\usepackage{verbatim}
\usepackage[linkcolor=black,citecolor=black,urlcolor=black,colorlinks=true]{hyperref}

\graphicspath{{figures/}}
\DeclareGraphicsExtensions{.png,.jpg,.eps,.pdf}
\IEEEoverridecommandlockouts
\title{\LARGE \bf VINS-Multi: A Robust Asynchronous Multi-camera-IMU State Estimator}
\author{Luqi Wang, Yang Xu and Shaojie Shen
\thanks{All authors are with the Department of Electronic and Computer Engineering, Hong Kong University of Science and Technology, Hong Kong, China. {\tt\footnotesize $\{$lwangax, yxuew$\}$@connect.ust.hk, eeshaojie@ust.hk}.}%
}
\begin{document}

\maketitle
\thispagestyle{empty}
\pagestyle{empty}

\begin{abstract}

State estimation is a critical foundational module in robotics applications, where robustness and performance are paramount. Although in recent years, many works have been focusing on improving one of the most widely adopted state estimation methods, visual inertial odometry (VIO), by incorporating multiple cameras, these efforts predominantly address synchronous camera systems. Asynchronous cameras, which offer simpler hardware configurations and enhanced resilience, have been largely overlooked. To fill this gap, this paper presents VINS-Multi, a novel multi-camera-IMU state estimator for asynchronous cameras. The estimator comprises parallel front ends, a front end coordinator, and a back end optimization module capable of  handling asynchronous input frames. It utilizes the frames effectively through a dynamic feature number allocation and a frame priority coordination strategy. The proposed estimator is integrated into a customized quadrotor platform and tested in multiple realistic and challenging scenarios to validate its practicality. Additionally, comprehensive benchmark results are provided to showcase the robustness and superior performance of the proposed estimator.

\end{abstract}

\section{Introduction}
\label{sec:introduction}

State estimation serves as a fundamental component within robotic applications, underpinning higher-level tasks with essential support. Therefore, the robustness and concision of the system are required during practice. A lot of recent research has concentrated on enhancing the performance and resilience of visual-inertial odometry (VIO), one of the predominant methods utilized for state estimation \cite{campos2021orb, leutenegger2015keyframe, qin2017vins}.

An intuitive and effective scheme for enhancing VIO involves the incorporation of additional cameras to cover different directions, thereby acquiring more information from the surrounding environment \cite{zhang2021balancing,jaekel2020robust,he2022towards}. Most of the works primarily investigated the use of synchronous cameras, which necessitate auxiliary hardware triggering mechanisms. However, practically, numerous cameras lack the hardware synchronization capability and systems reliant on synchronous cameras often succumb to malfunctions induced by faulty trigger signals or partial camera failures, which stem from design or technical issues. Hence, during some real applications in demanding environments, for instance wilderness and construction sites, deploying multiple asynchronous cameras can be less cumbersome and offer superior robustness against failures. Although \cite{yang2021asynchronous} has explored the use of multiple asynchronous cameras on a driving dataset, the approach combines asynchronous frames as multi-frame batches and approximates the states by a B-spline trajectory for interpolation, which typically loses degrees of freedom. To circumvent this limitation, the incorporation of an inertial measurement unit (IMU), a commonly adopted sensor in robotics, can be a simple and effective solution. As a result, we propose the VINS-Multi, an asynchronous multi-camera-IMU state estimator developed from our previous work \cite{qin2017vins,qin2019a}.

\begin{figure}[t]
\begin{center}
\subfigure[\label{fig:tunnel_vent_1} Vent pipe case 1.]
{\includegraphics[height=0.3\columnwidth]{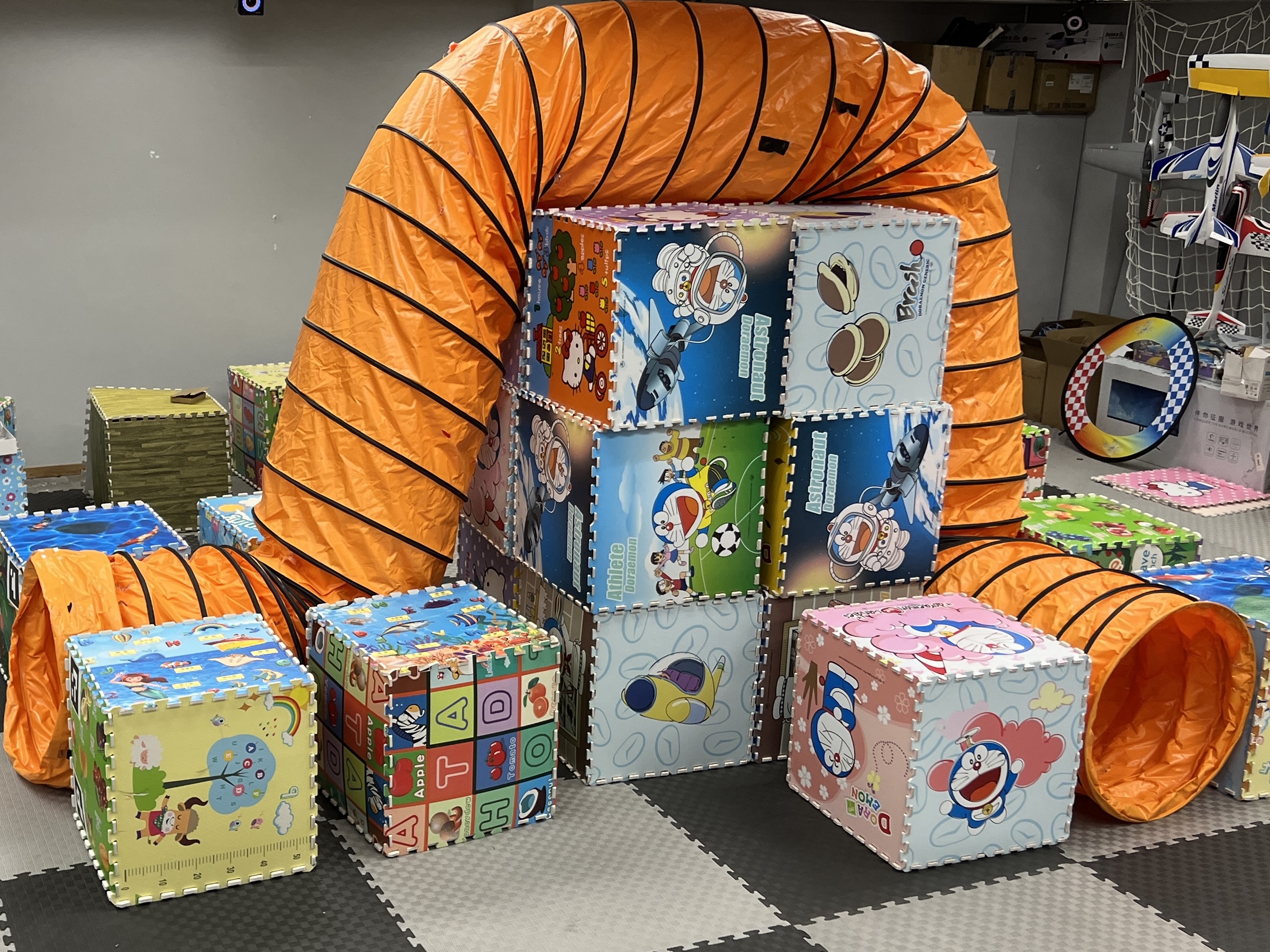}} 
\subfigure[\label{fig:vent_1_rviz} Visualization of vent pipe case 1.]
{\includegraphics[height=0.3\columnwidth]{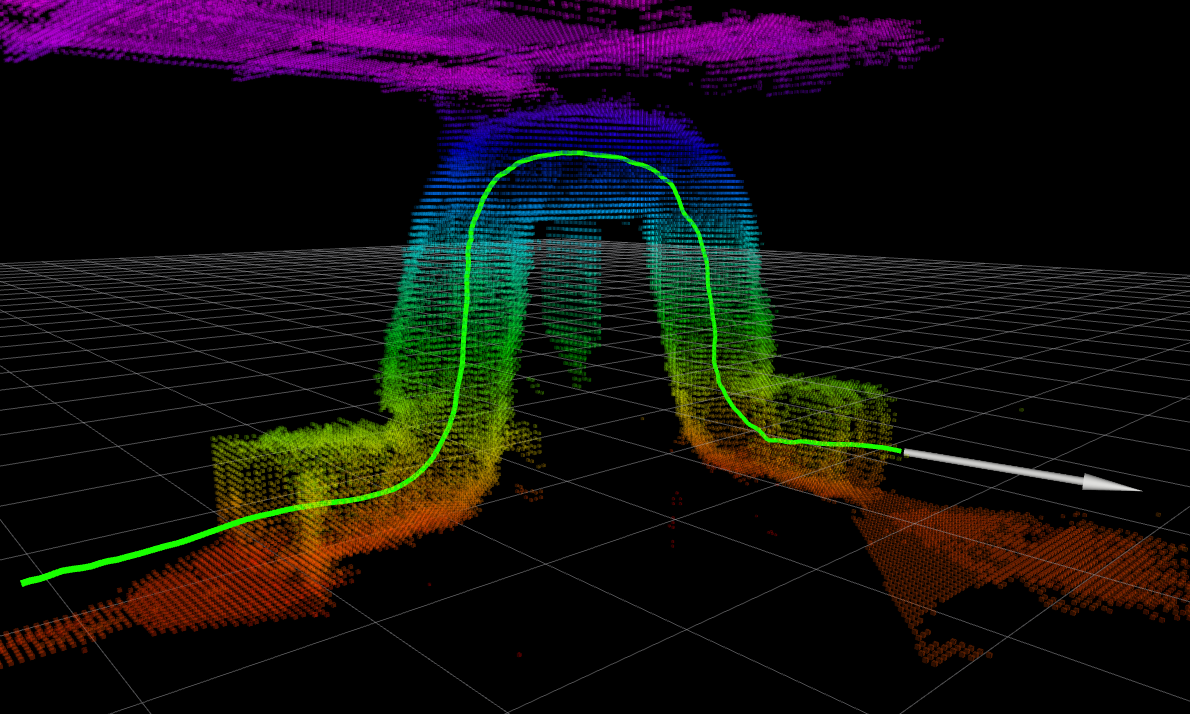}} 
\subfigure[\label{fig:tunnel_vent_2} Vent pipe case 2.]
{\includegraphics[height=0.3\columnwidth]{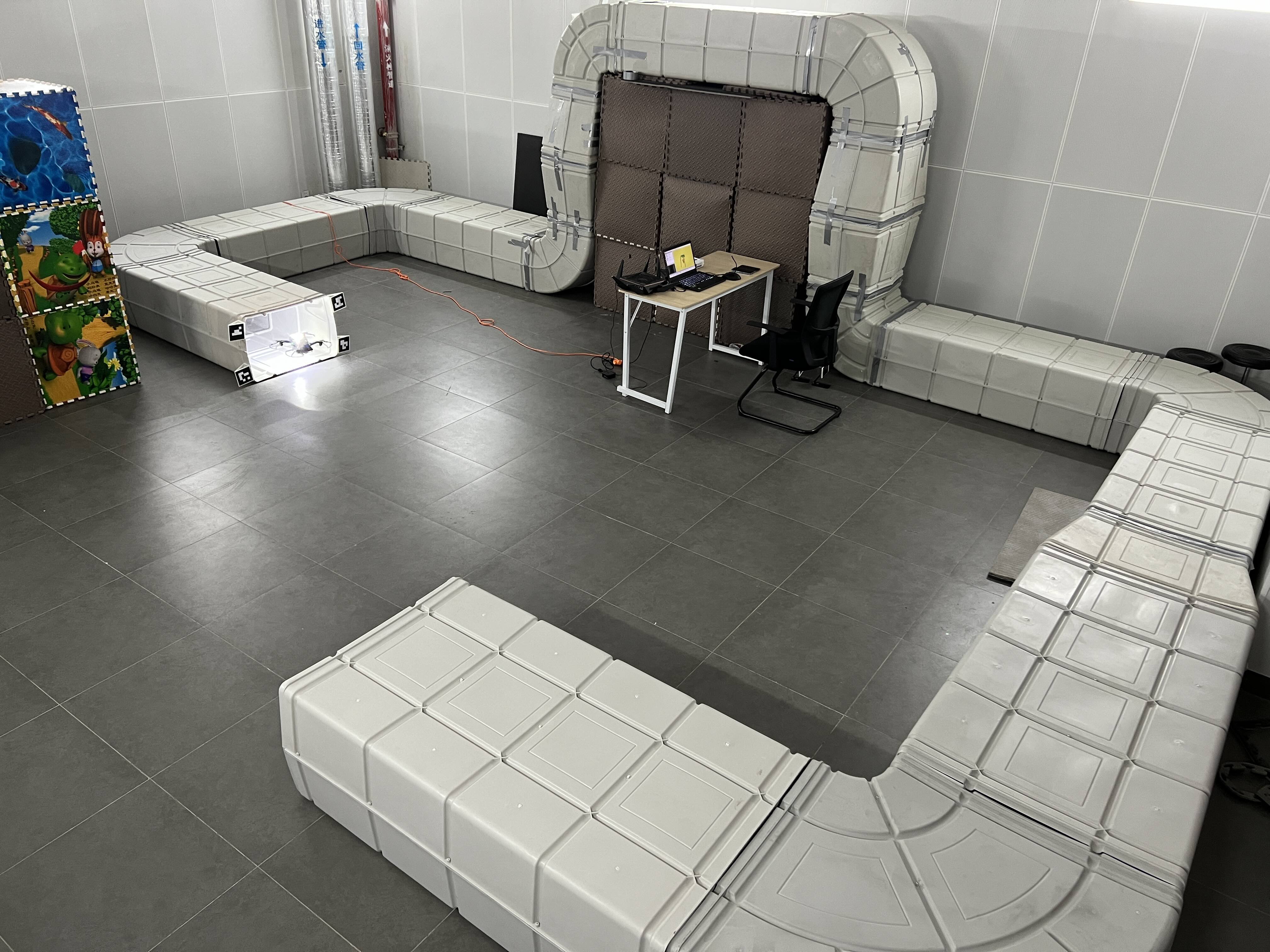}} 
\subfigure[\label{fig:vent_2_rviz} Visualization of vent pipe case 2.]
{\includegraphics[height=0.3\columnwidth]{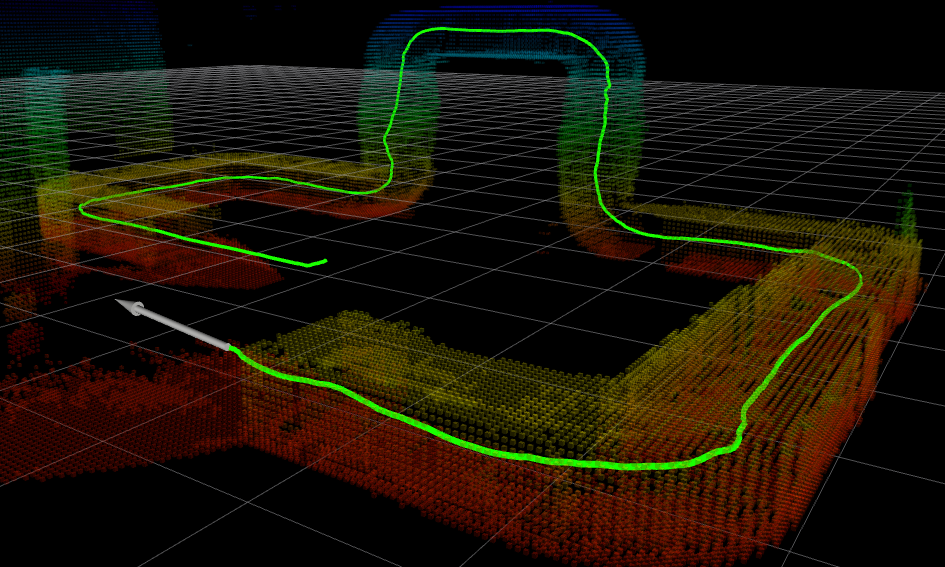}} 
\end{center}
\vspace{-0.3cm}
\caption{\label{fig:tunnel} The deployment of VINS-Multi in aerial vent pipe inspection scenarios. The green lines are the estimated trajectory and the white arrows indicate the estimated odometries. The maps are constructed during the flights and the color code indicates the height.}
\vspace{-1.2cm}
\end{figure}

The contributions of this paper are the following:
\begin{enumerate}
	\item We design a novel dynamic feature number allocation alongside a frame priority coordination strategy to efficiently handle the asynchronous frame inputs.
	\item We combine a parallel front end and a front end coordinator based on the proposed strategy, as well as a sliding window optimization module into a robust multi-camera-IMU state estimator that can accommodate multiple asynchronous cameras of mixed types.
	\item The estimator is integrated into a quadrotor platform for extensive experiments in realistic and challenging scenarios, and the benchmark results are presented.
\end{enumerate}

\begin{figure}[t]
\begin{center}
{\includegraphics[width=0.8\columnwidth]{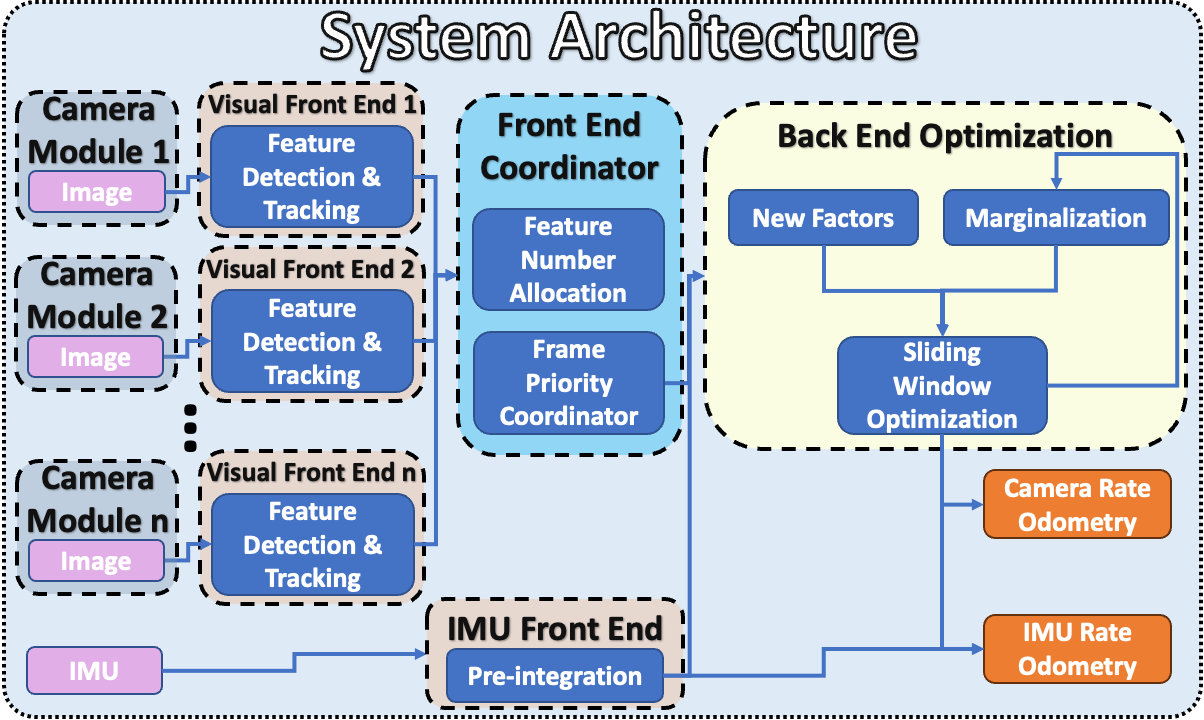}} 
\end{center}
\vspace{-0.4cm}
\caption{\label{fig:system} The system architecture of the proposed state estimator. Note that the camera modules can be of mixed types.}
\vspace{-0.4cm}
\end{figure}

\section{Methodology}
\label{sec:methodology}

The entire system architecture is depicted in Fig. \ref{fig:system}. The images, accompanied by depth if using RGBD cameras, are sent to the visual front ends and the processed measurements are subsequently coordinated by the front end coordinator. The coordinator dynamically redistributes the number of features processed by each front end based on the collected features, as well as determines whether the measurements are forwarded to the sliding window in the back end optimization module according to the priority. The back end obtains the visual and depth measurements, as well as the pre-integration results from the raw IMU input to output the camera and IMU rate odometries at 30 and 500 Hz after optimization.

\subsection{Front End}
\label{subsec:front_end}
The visual front end adheres to a standard procedure which includes feature detection, tracking and outlier rejection, similar to the methodologies established by VINS-Mono \cite{qin2017vins} and VINS-Fusion \cite{qin2019a}, while the front ends are running in parallel with each front end thread handling a separate camera module. The extracted visual measurements, along with the corresponding depth measurements (when using RGBD cameras), are firstly sent to the front end coordinator preceding the back end optimization. The IMU front end performs pre-integration and directly outputs the high rate odometry based on the latest optimization results.

\subsection{Front End Coordinator}
\label{subsec:front_end_coordinator}

\subsubsection{Feature Number Allocation}
\label{subsubsec:feature_number_update}

Considering the different feature qualities caused by the varying captured scenes from cameras facing distinct directions
, it is inefficient to treat the cameras equally. To make effective use of the computation resources, the maximum extracted feature number from images of each camera is dynamically allocated according to the scene. In particular, given the total maximum feature number $FN$ we allocate $fn_{i}$ features for camera $i$ according to the tracked feature number from the last frame $tfn_{i}$:
\begin{equation}\label{eq:feature_alloc}
	fn_{i} =\frac{tfn_{i}}{\sum\limits_{i=0}^{N} tfn_{i}} FN.
\end{equation}
Note that by dynamically adjusting the maximum feature number of each camera, the feature tracking rate will converge to the same across the cameras, meaning that when a front end has high feature tracking rate, larger maximum feature number will be allocated to it.

\subsubsection{Frame Priority Coordinator}
\label{subsubsec:frame_priority}
Since in some scenarios, not all the frames are necessary to be inserted into the sliding window for optimization, for instance when a camera is obstructed, or the time interval between two frames is excessive compared with others, a frame priority coordinator is required to handle the priority among the cameras so as to make efficient use of the computation resources. The priority of a frame of camera $i$ depends on both the feature and time interval $\delta t_i$ since the last frame from camera $i$ was received. The feature priority $P_{fi}$ is determined by the ratio between the current maximum feature number of camera $i$ $fn_{i}$ and the total maximum feature number $FN$ :
\begin{equation}\label{eq:feature_priority}
	P_{fi} =\frac{fn_{i}}{FN},
\end{equation}
and the frame time interval priority $P_{ti}$ is formulated as: 
\begin{equation}\label{eq:time_priority}
	P_{ti} = e^{- k \delta t_i},
\end{equation}
where $k$ is a positive constant.
The coordinator waits for the frame with either the highest feature priority to maintain consistent feature tracking quality or the highest time priority to preclude camera failures and forwards the corresponding measurements to the back end for optimization.

\subsection{Back End Optimization}
\label{subsec:back_end}

\begin{figure}[t]
\begin{center}
{\includegraphics[width=0.61\columnwidth]{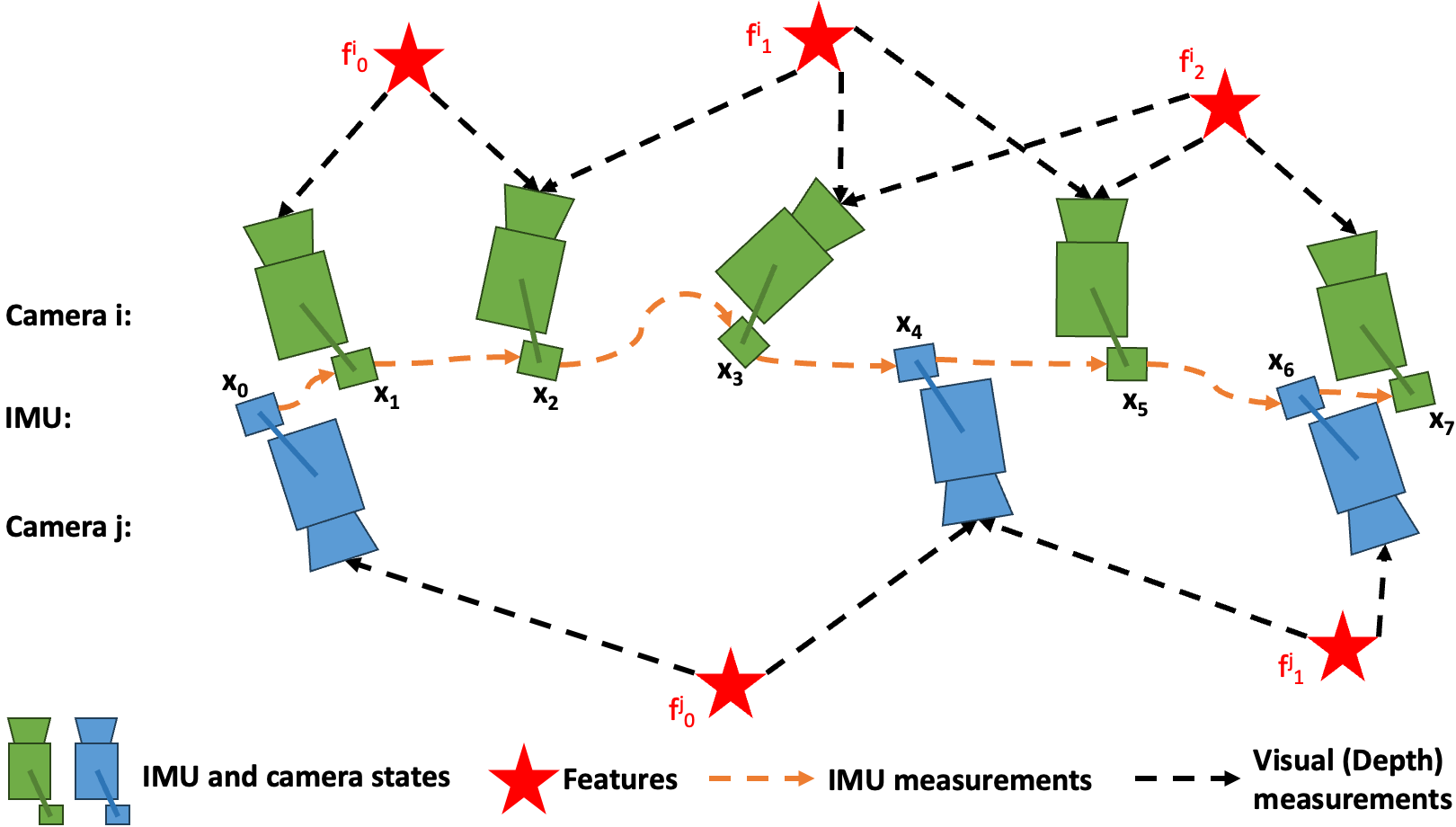}} 
\end{center}
\vspace{-0.5cm}
\caption{\label{fig:sliding_window} A schematic of the sliding window incorporating IMU data and visual (including depth) data for optimization.}
\vspace{-1.0cm}
\end{figure}

\begin{figure}[t]
\begin{center}
{\includegraphics[width=1.0\columnwidth]{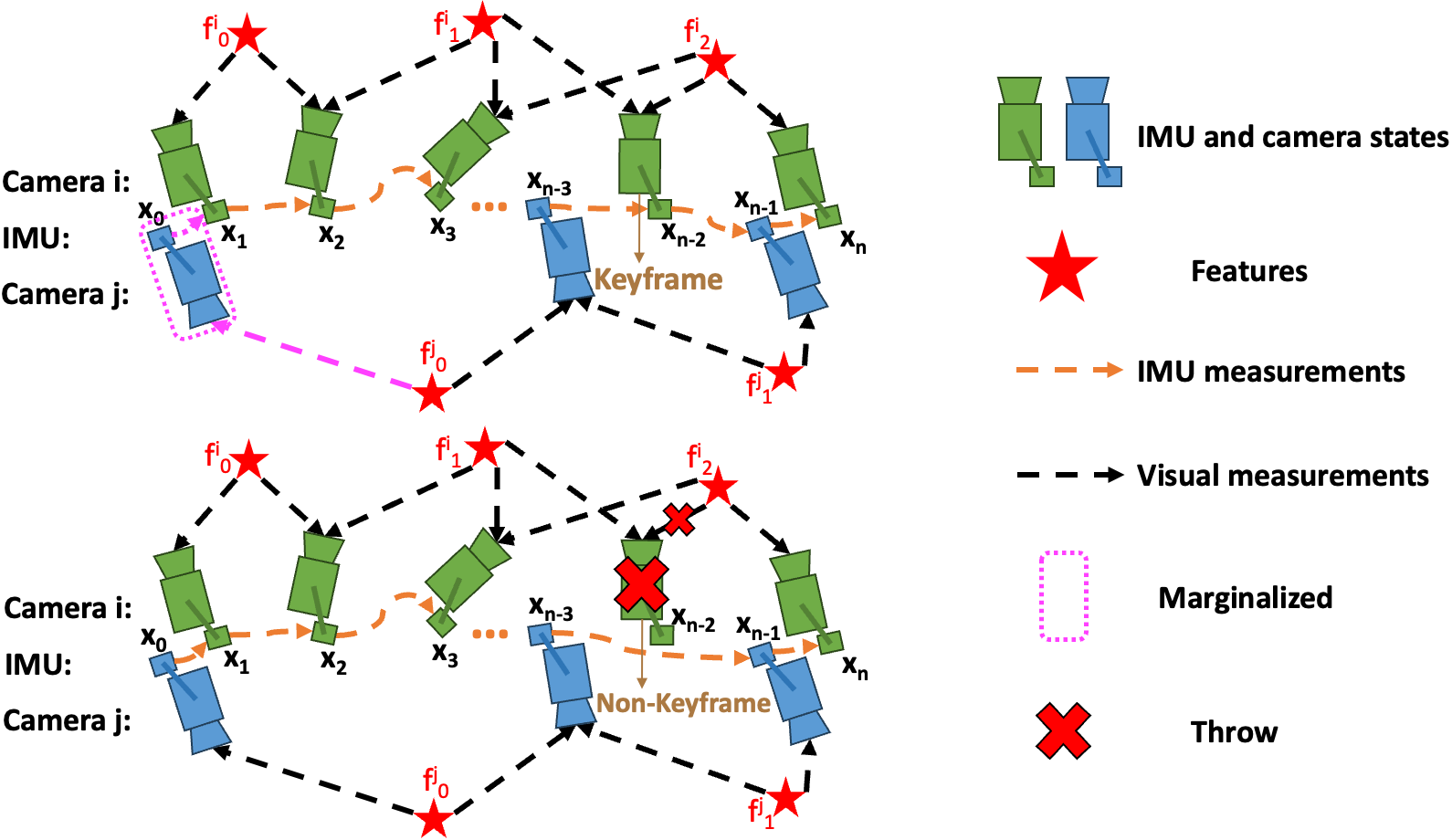}} 
\end{center}
\vspace{-0.4cm}
\caption{\label{fig:marginalization} An illustration of the marginalization strategy.}
\vspace{-0.2cm}
\end{figure}

In the back end optimization, we extend the formulation from our previous work \cite{qin2017vins, qin2019a} to accommodate multiple asynchronous cameras. As depicted in Fig. \ref{fig:sliding_window}, the accepted frames from the cameras are chronologically ordered inside the sliding window according to their time stamps for optimization with a full state vector $\mathcal{X}$ defined as:

\begin{equation}\label{eq:var}
\begin{aligned}
\mathcal{X} & =\left[\mathbf{x}_0, \mathbf{x}_1, \cdots \mathbf{x}_n, \lambda_0, \lambda_1, \cdots \lambda_l, \mathbf{x}_{c_0}, \mathbf{x}_{c_1}, \cdots, \mathbf{x}_{c_{N-1}} \right] \\
\mathbf{x}_i & =\left[\mathbf{p}_i^w, \mathbf{v}_i^w, \mathbf{R}_i^w, \mathbf{b}_a, \mathbf{b}_g\right], i \in[0, n] \\
\mathbf{x}^{c}_{k} &= \left[ \mathbf{p}_{c_{k}}^b, \mathbf{R}_{c_{k}}^b, t_{d_{k}} \right], k \in[0, N-1],
\end{aligned}
\end{equation}
where $\mathbf{x}_i$ is the $i$-th IMU and camera states consists of position, velocity, rotation as well as acceleration and gyroscope biases, $\lambda_j$ denotes the inverse depth of the $j$-th feature upon its initial observation, and $\mathbf{x}^{c}_{k}$ encompasses of the extrinsic translation, rotation and the time offset relative to the IMU of the $k$-th camera.
The optimization objective is formulated similarly as the combination of the prior factor, the IMU propagation factor and the visual (depth) factor:

\begin{equation}\label{eq:opt_obj}
\begin{aligned}
\min _{\mathcal{X}}\{\underbrace{\left\|\mathbf{e}_p-\mathbf{H}_p \mathcal{X}\right\|^2}_{\text {prior factor }} & +\underbrace{\sum_{k \in \mathcal{B}}\left\|\mathbf{e}_{\mathcal{B}}\left(\mathbf{z}_{k+1}^k, \mathcal{X}\right)\right\|_{\mathbf{P}_{k+1}^k}^2}_{\text {IMU propagation factor }} \\
& +\underbrace{\sum_{(l, j) \in \mathcal{C}}\left\|\mathbf{e}_{\mathcal{C}}\left(\mathbf{z}_l^j, \mathcal{X}\right)\right\|_{\mathbf{P}_l^j}^2}_{\text {visual (depth) factor }}\} .
\end{aligned}
\end{equation}
The detailed formulation can be found in \cite{qin2017vins}, while a depth (re-projection) error is integrated to facilitate RGBD cameras. The prior factor originates from the marginalization of the frames, which follows a different strategy from preceding research, as illustrated in Fig. \ref{fig:marginalization}. Upon the arrival of a new frame from camera $i$, the last frame from camera $i$ is checked. If it is a keyframe, the most outdated frame in the sliding window is marginalized, regardless of the originating camera. Otherwise, the last frame from camera $i$ is thrown. This strategy is able to handle uneven frames from multiple cameras and cope with the malfunction of certain cameras.

\section{Experiment and Results}
\label{sec:exp}

\subsection{Experiment Setup}
\label{subsec:exp_setup}

\begin{figure}[t]
\begin{center}
\subfigure[\label{fig:drone} The custom-built quadrotor platform equipped with three RGBD cameras.]
{\includegraphics[width=0.72\columnwidth]{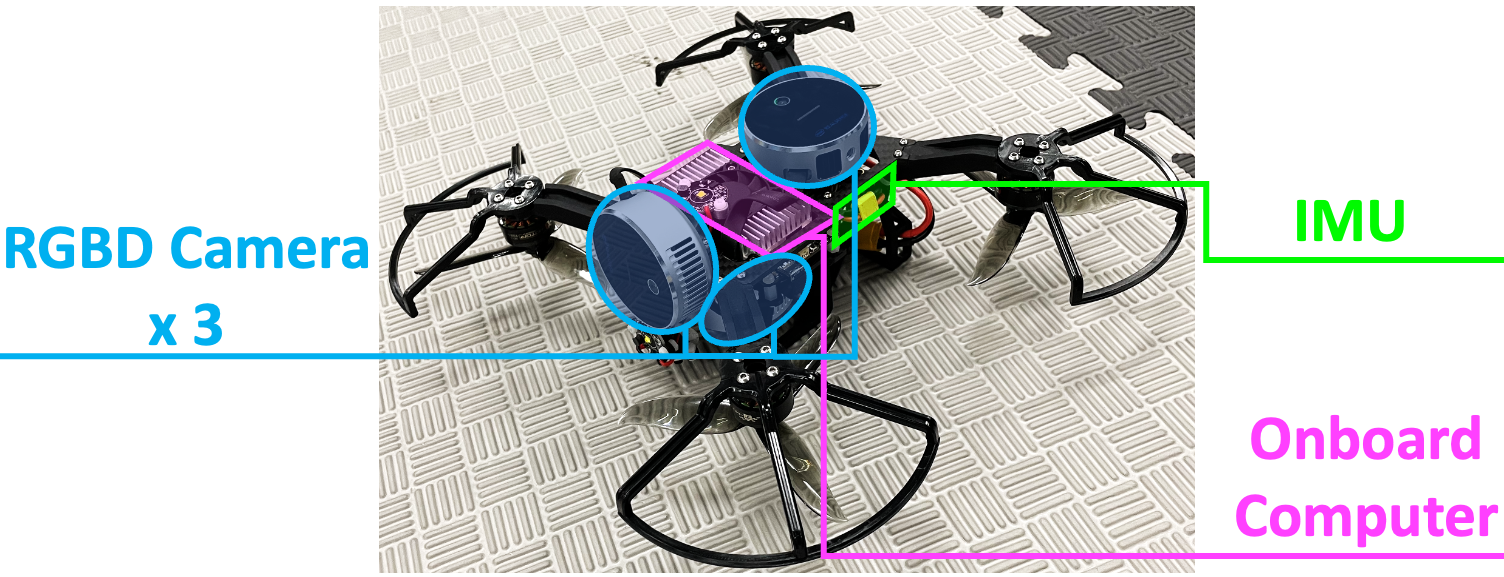}}             
\subfigure[\label{fig:drone_435} Replace the top RGBD camera by a stereo camera.]
{\includegraphics[width=0.8\columnwidth]{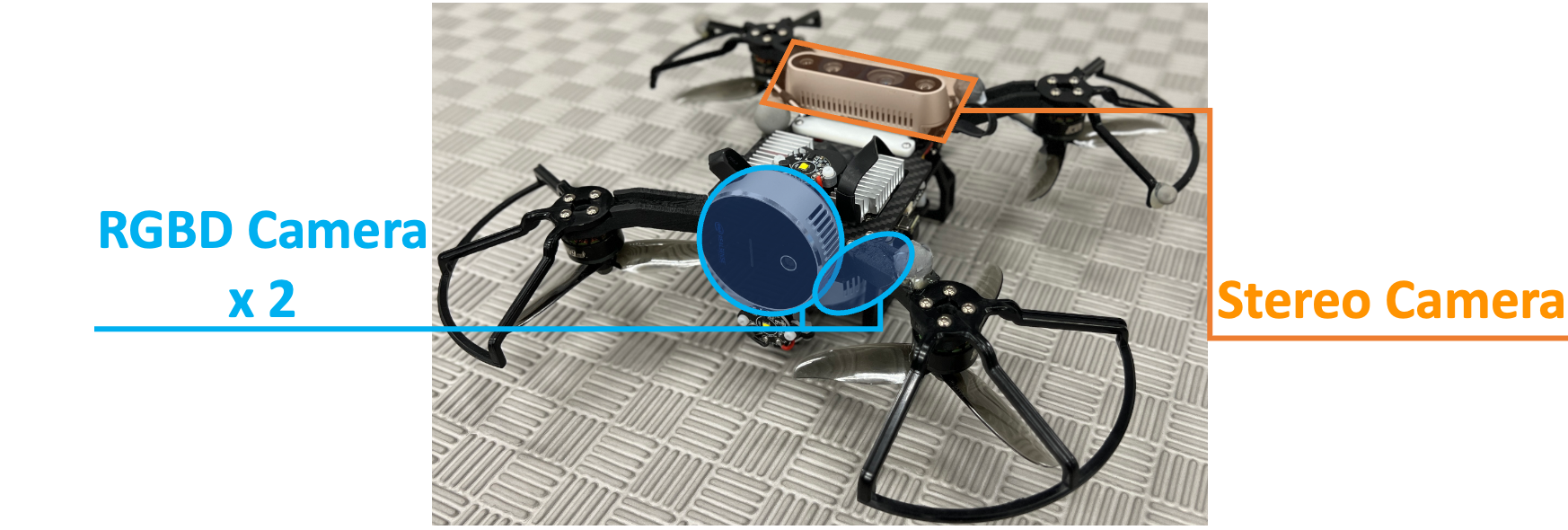}} 
\end{center}
\vspace{-0.2cm}
\caption{\label{fig:hard} The quadrotor deployed in experiments. The system is compatible with mixed types of monocular, RGBD, and stereo camera modules.}
\vspace{-2.2cm}
\end{figure}


\begin{figure}[t]
\begin{center}
\subfigure[\label{fig:plug_lid} Cover the top camera with a lid.]
{\includegraphics[width=0.48\columnwidth]{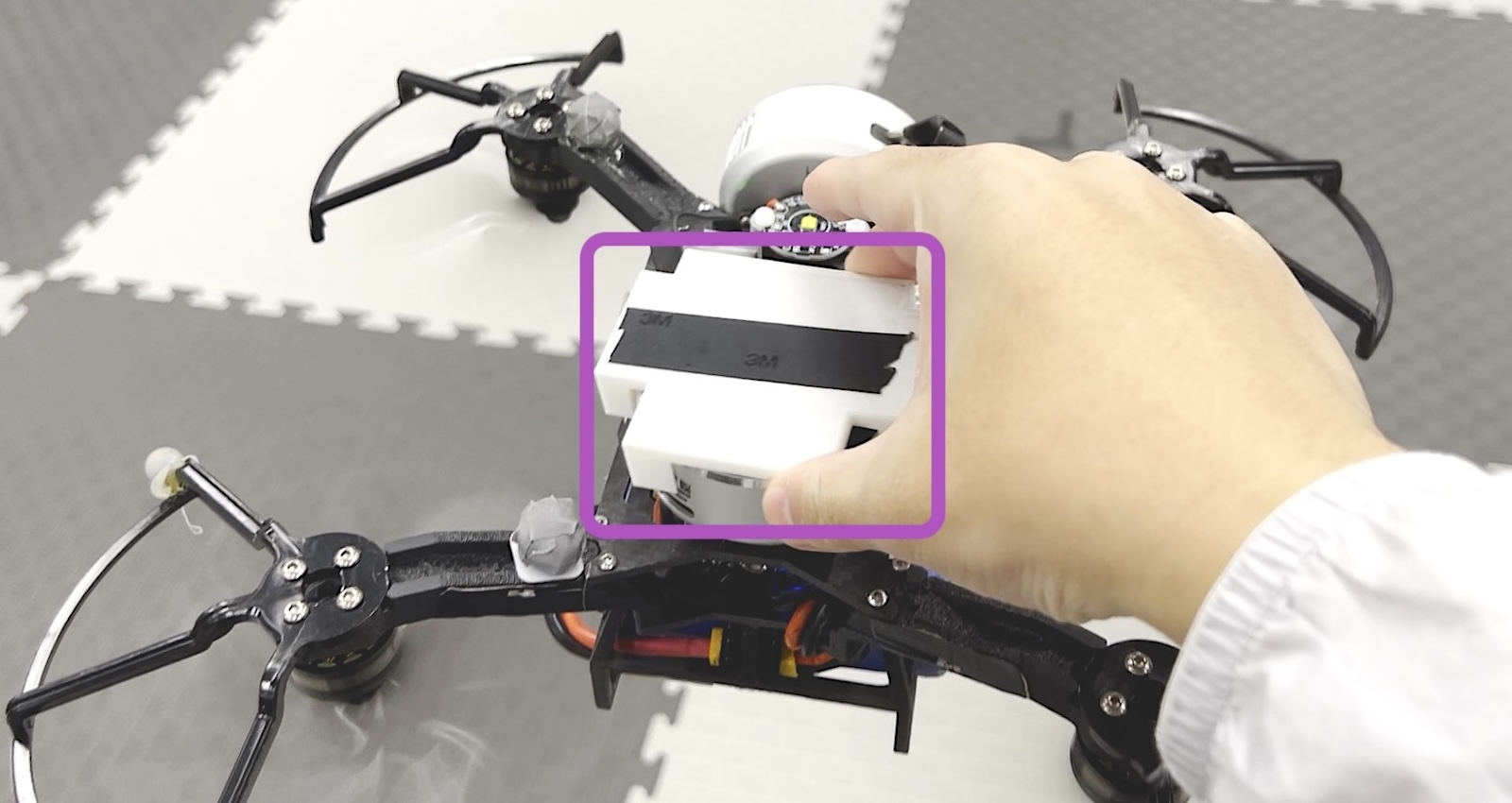}} 
\subfigure[\label{fig:plug_unplug} Unplug the top camera.]
{\includegraphics[width=0.48\columnwidth]{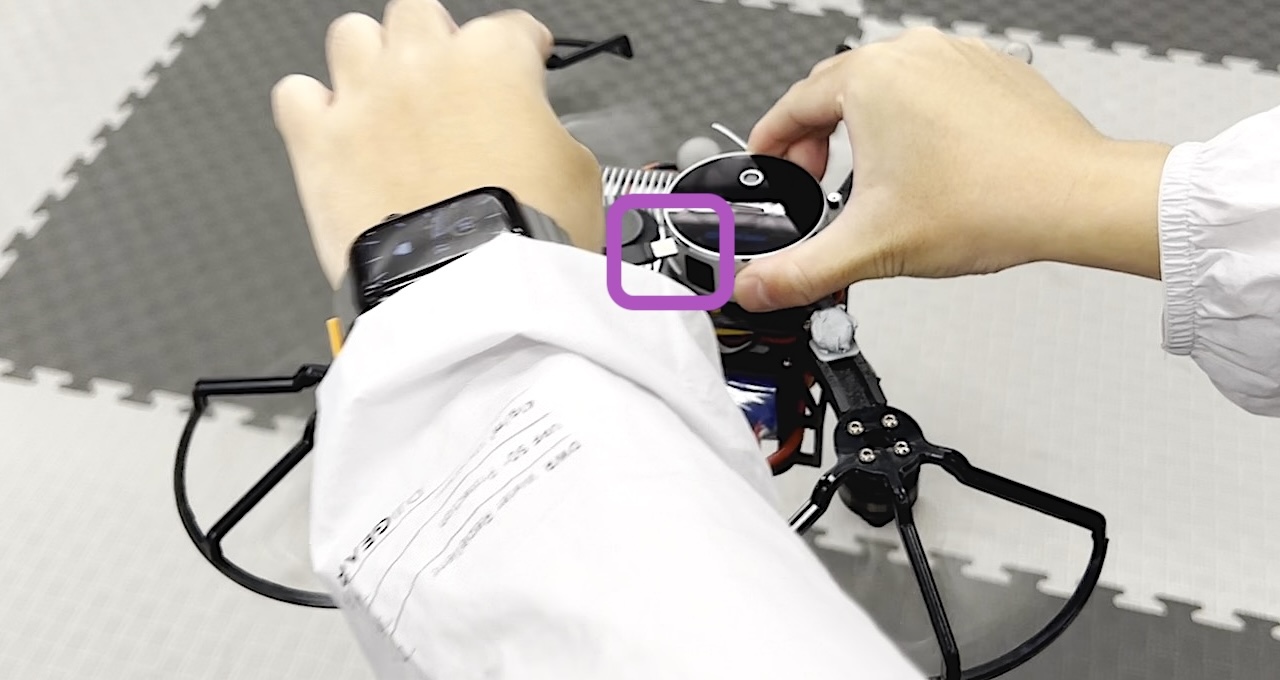}} 
\subfigure[\label{fig:plug_plug} Plug the front camera.]
{\includegraphics[width=0.48\columnwidth]{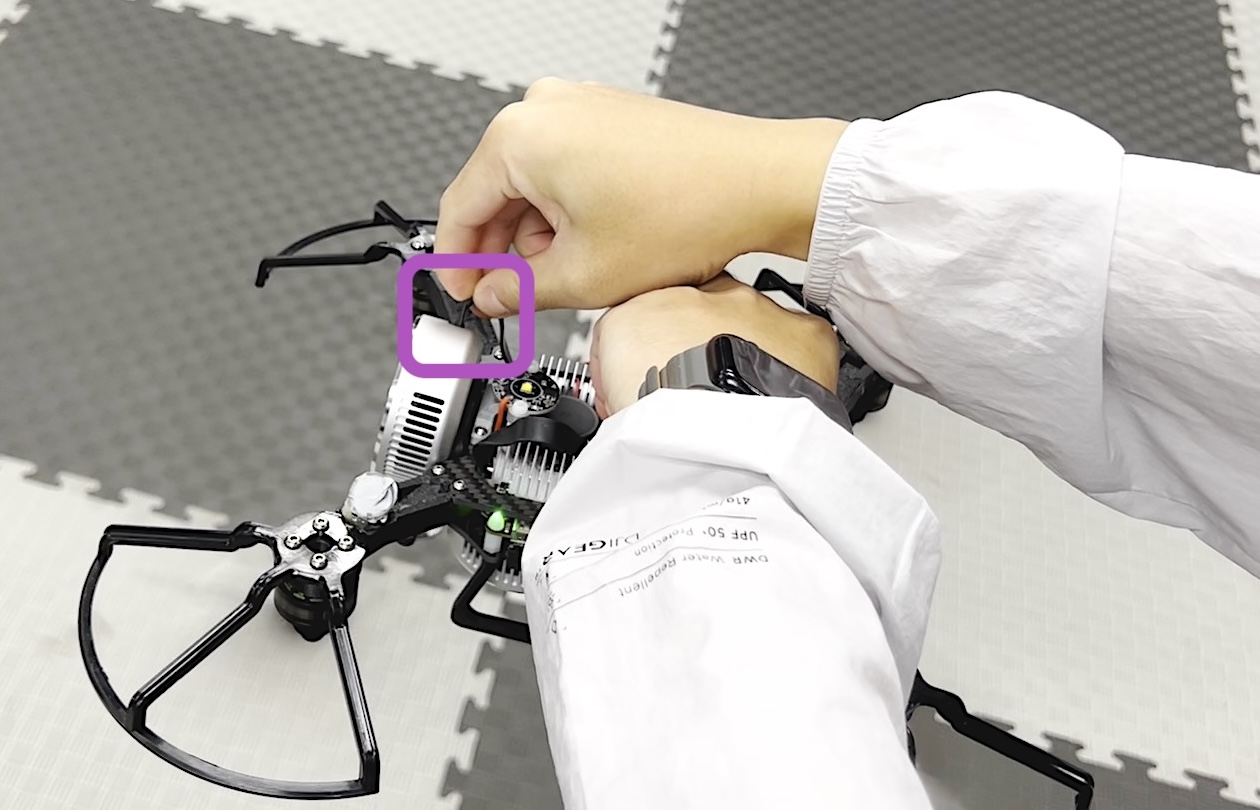}} 
\subfigure[\label{fig:plug_traj} Trajectory result.]
{\includegraphics[width=0.48\columnwidth]{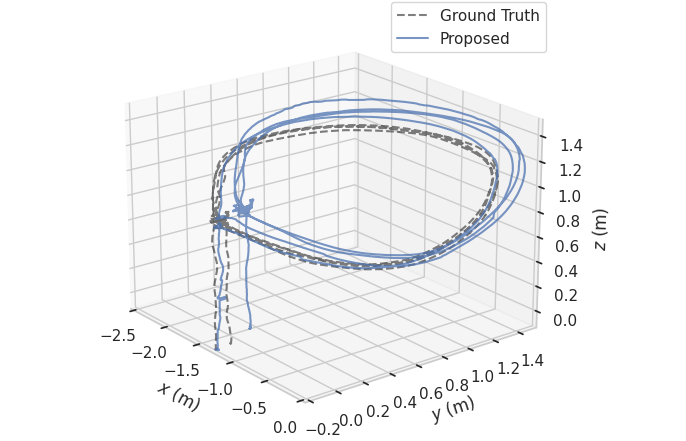}} 
\end{center}
\vspace{-0.4cm}
\caption{\label{fig:plug} The procedure and comparison of the trajectory results against the ground truth during the camera failure and recovery scenario.}
\vspace{-0.4cm}
\end{figure}

\begin{figure}[H]
\begin{center}
\subfigure[\label{fig:vins_wall} Snapshot of the scenario with scarce features for stable tracking in the marked areas.]
{\includegraphics[width=0.7\columnwidth]{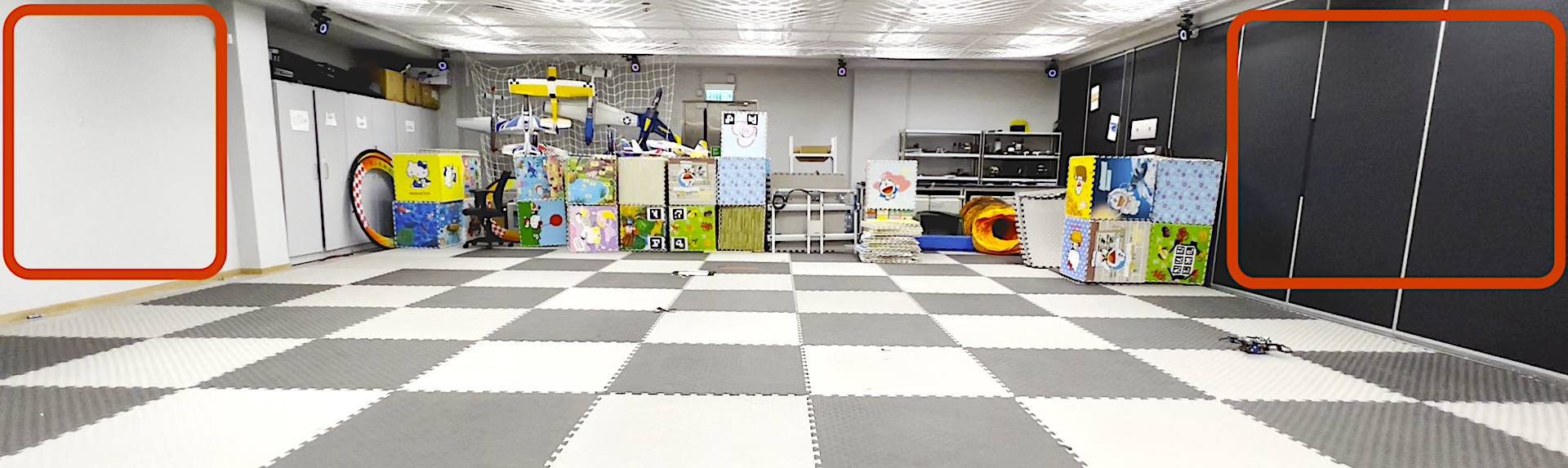}} 
\subfigure[\label{fig:wall_traj} Trajectory result.]
{\includegraphics[height=0.37\columnwidth]{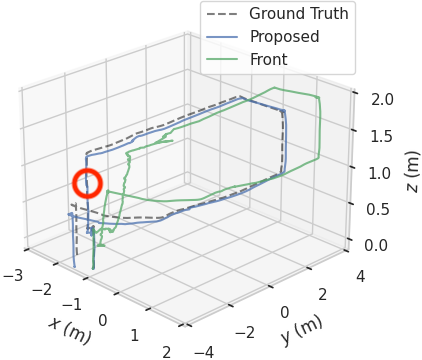}} 
\subfigure[\label{fig:wall_front_feature} Features from the front camera.]
{\includegraphics[height=0.37\columnwidth]{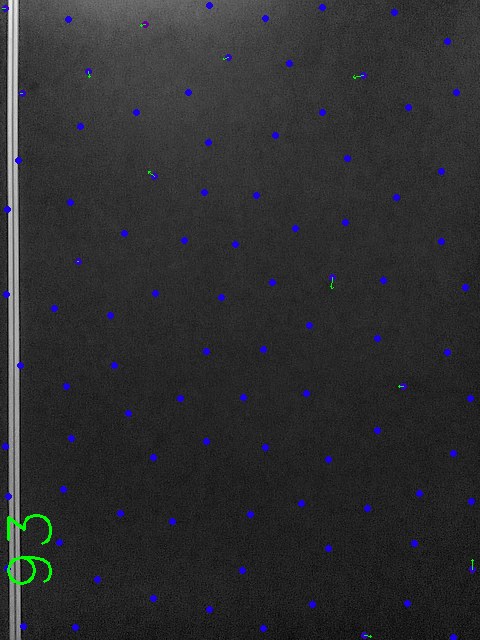}} 
\subfigure[\label{fig:wall_multi_feature} Features from three cameras using the proposed method.]
{\includegraphics[height=0.37\columnwidth]{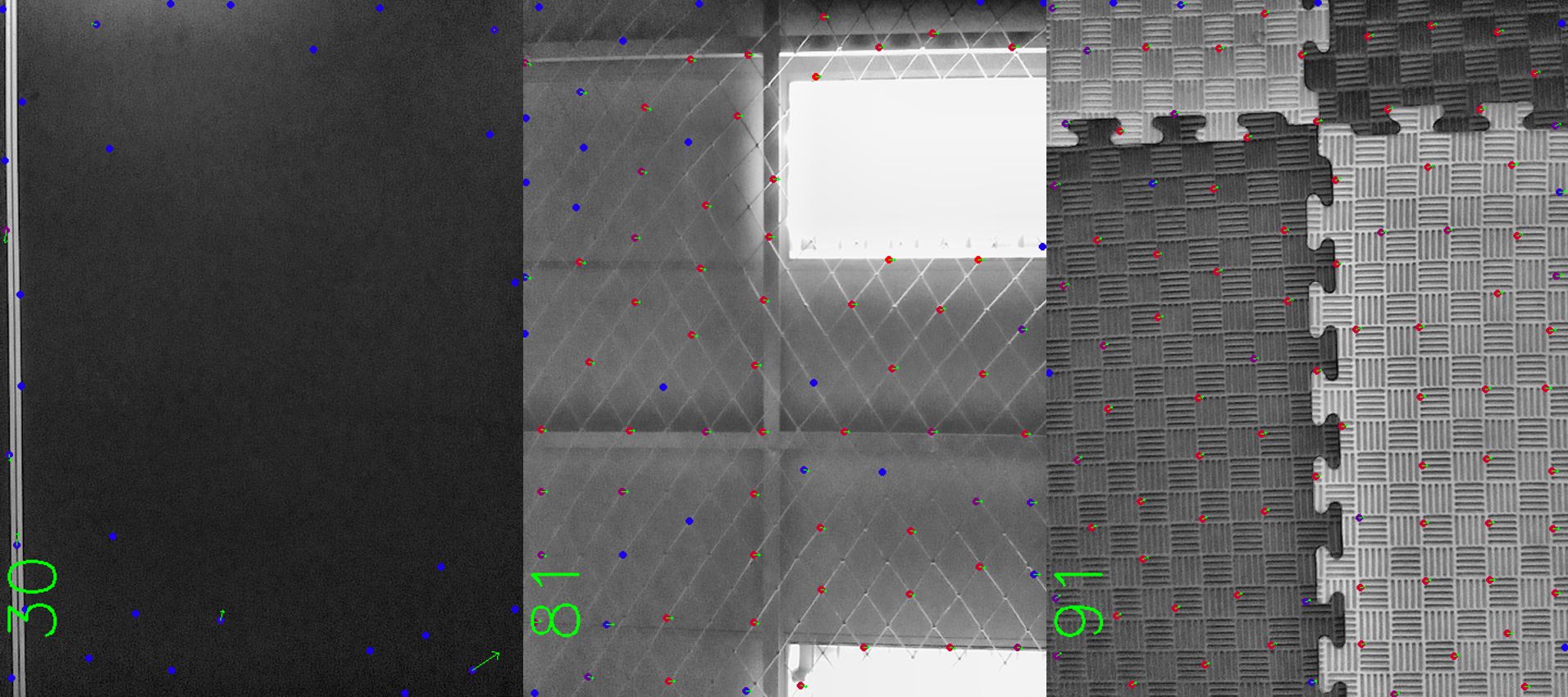}} 
\end{center}
\vspace{-0.4cm}
\caption{\label{fig:wall_result} The scene and the comparison of the trajectory results of the proposed method using three cameras and using solely the front camera with the ground truth and the features extracted at the circled position from the front, top, and bottom images in the wall inspection scenario. The quantity of allocated features is indicated on the lower left of each image, with the feature point color gradient representing tracking duration from blue (shortest) to red (longest).}
\vspace{-0.2cm}
\end{figure}

The evaluation of the proposed state estimator is conducted using a custom-built quadrotor platform depicted in Fig. \ref{fig:hard}. The quadrotor is equipped with three Intel Realsense L515 RGBD cameras, while in particular experiments, the top RGBD camera is replaced with an Intel Realsense D435 stereo camera to verify the performance on mixed types of cameras. The data from a BMI088 IMU embedded in the NxtPX4 flight controller is adopted throughout the test flights. All the computations are performed on an Nvidia Jetson Orin NX onboard computer. The experiments are conducted in four scenarios: camera failure and recovery shown in Fig. \ref{fig:plug}, wall inspection shown in Fig. \ref{fig:vins_wall}, flights employing mixed types of camera modules on the quadrotor in Fig. \ref{fig:drone_435}, and vent pipe inspection shown in Fig. \ref{fig:tunnel}.

\begin{table*}[h]
\centering
\caption{\label{tab:traj_benchmark}Benchmark results on the estimated trajectories}
\begin{tabular}{@{}lcccccccccc@{}}
\toprule
\multirow{4}{*}{Seq.} & \multicolumn{4}{c}{Multiple cameras} & \multicolumn{6}{c}{Single camera} \\
\cmidrule(r){2-5} \cmidrule(l){6-11}
& \multicolumn{2}{c}{Proposed} & \multicolumn{2}{c}{W/O feature allocation} & \multicolumn{2}{c}{Front} & \multicolumn{2}{c}{Top} & \multicolumn{2}{c}{Down} \\
\cmidrule(r){2-3} \cmidrule(r){4-5} \cmidrule(r){6-7} \cmidrule(r){8-9} \cmidrule(r){10-11}
& ATE(m) & RPE(m) & ATE(m) & RPE(m) & APE(m) & RPE(m) & ATE(m) & RPE(m) & ATE(m) & RPE(m) \\
\midrule
failure \& recovery  & \textbf{0.07711}  & \textbf{0.00042} & 2.45813 & 0.00160 & $\times$ & $\times$ & $\times$ & $\times$ & $\times$ & $\times$ \\
wall inspection  & \textbf{0.08488} & \textbf{0.00070}   & 0.28015  & 0.00077 & 0.44229 & 0.00284 & 3.57276   & 0.00065    & $\times$ & $\times$ \\
mixed camera types & \textbf{0.07596} & \textbf{0.00076} & 2.02272 & 0.00306 & 0.17287 & 0.00094 & 0.11617 & 0.00102 & $\times$ & $\times$ \\
\bottomrule
\multicolumn{2}{l}{$\times$: fail. \textbf{Bold}: best results.}
\end{tabular}
\vspace{-0.4cm}
\end{table*}

\subsection{Result and Analysis}
\label{sec:result}

\begin{figure*}[t]
\begin{center}
\subfigure[\label{fig:435_traj} Trajectory results.]
{\includegraphics[height=0.37\columnwidth]{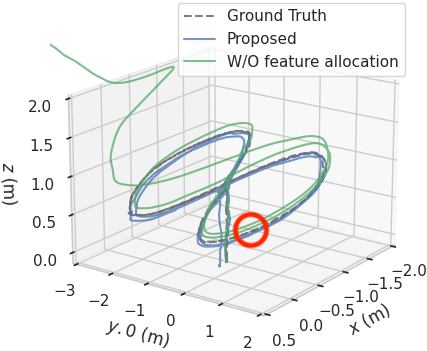}}
\subfigure[\label{fig:435_multi_avg_feature} The features without feature allocation.]
{\includegraphics[height=0.37\columnwidth]{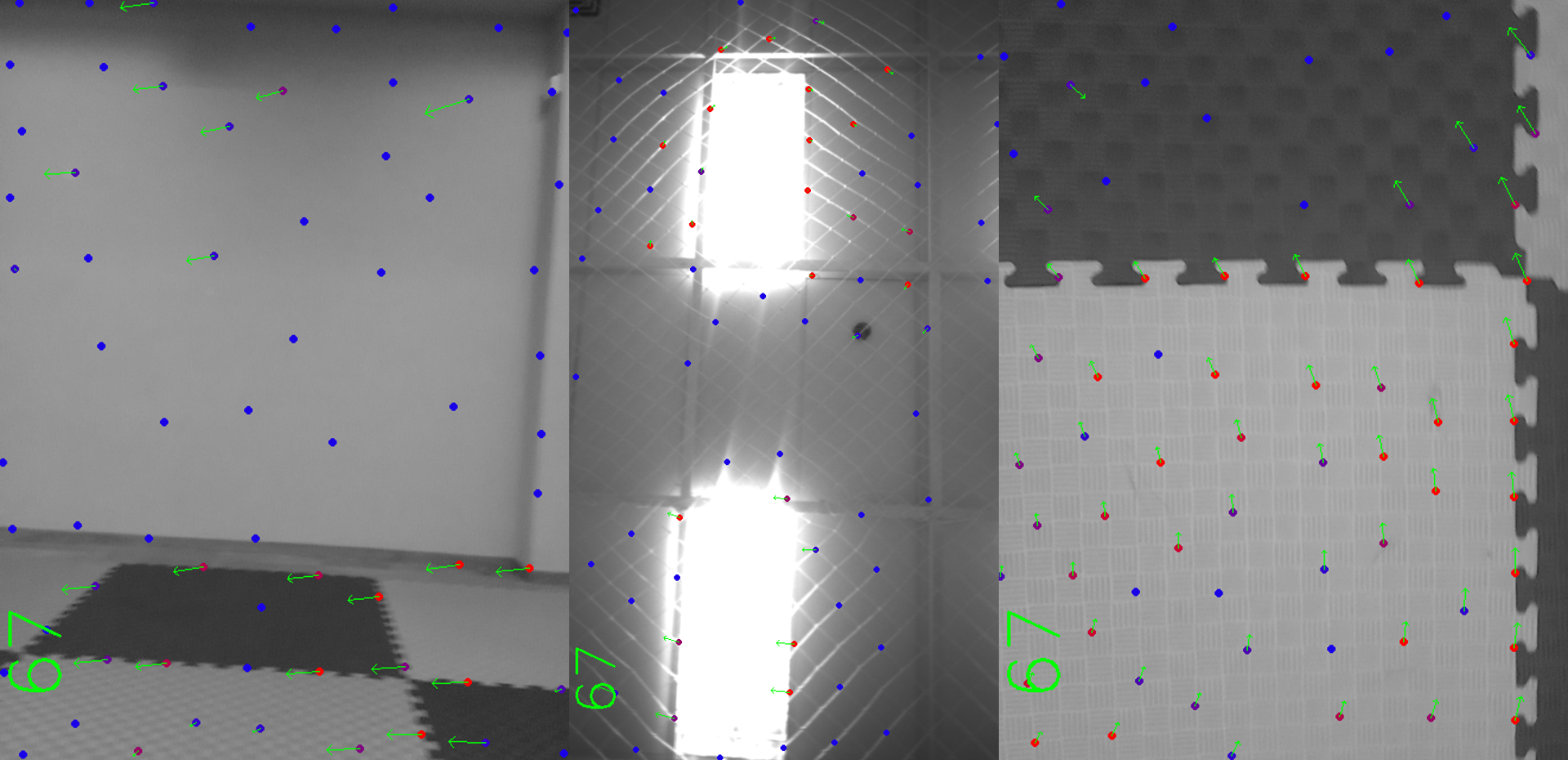}} 
\subfigure[\label{fig:435_multi_feature} The features using the proposed method.]
{\includegraphics[height=0.37\columnwidth]{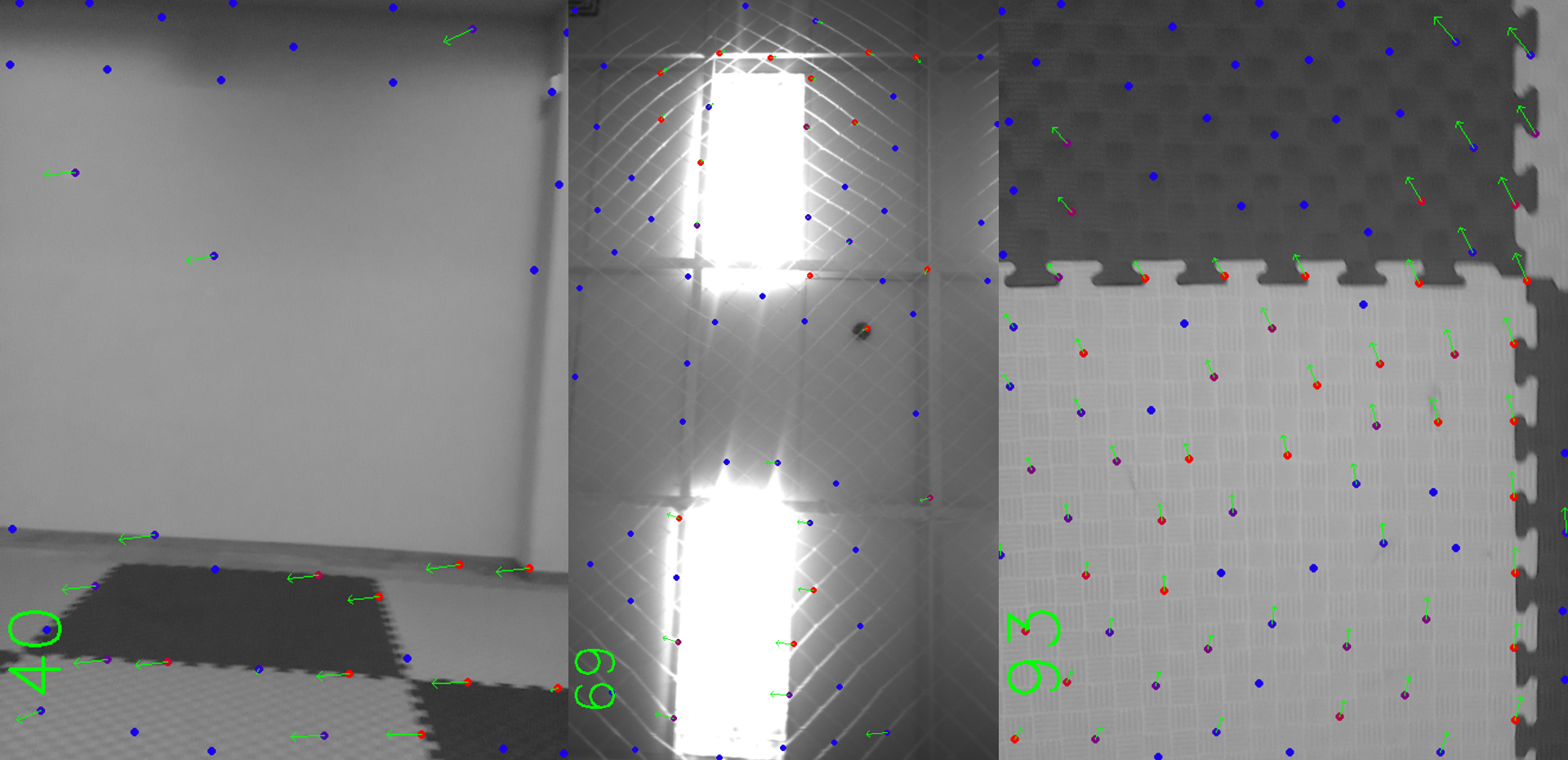}} 
\end{center}
\vspace{-0.4cm}
\caption{\label{fig:435_result} The comparison of the trajectory results of the proposed method and the method without feature allocation with the ground truth and the features extracted at the circled position on the front, top, and bottom images in the mixed camera types configuration scenario. The quantity of allocated features is indicated on the lower left of each image, with the feature point color gradient representing tracking duration from blue (shortest) to red (longest).}
\vspace{-0.4cm}
\end{figure*}

During the failure and recovery scenario, the quadrotor first takes off with only its top and bottom cameras. The robustness of the system is tested through a sequence of procedures consisting of covering the top camera with a lid, removing the lid, unplugging the top camera, and finally plugging the front camera during the flight. As evidenced by Fig. \ref{fig:plug_traj}, the proposed system maintains functionality amid camera failures, exhibiting decent estimation errors, and the recovered front camera can be successfully added into the estimation, demonstrating the superior robustness, which is unattainable by adopting single cameras.

The trajectory result and comparison with a solitary front camera during the wall inspection scenario is shown in Fig. \ref{fig:wall_traj}. When the single front camera faces the black wall during take-off, we notice a conspicuous trajectory drift of the estimated trajectory due to the lack of stable tracking features (see Fig. \ref{fig:wall_front_feature}), while the proposed method using multiple cameras is capable of handling this case with stable feature tracking using the top and bottom cameras (as shown in Fig. \ref{fig:wall_multi_feature}).

The trajectory result and comparison with the ablation of dynamic feature allocation in flights with mixed camera types configuration is shown in Fig. \ref{fig:435_traj}. The method lacking dynamic feature allocation is prone to substantial drift during rapid yaw movements, particularly apparent when wasting unnecessary features on feature-poor walls (see Fig. \ref{fig:435_multi_avg_feature}). Conversely, the proposed method copes effectively with this challenging scenario  buttressed by its dynamic feature allocation strategy, as shown in Fig. \ref{fig:435_multi_feature}.

The efficacy and robustness of the proposed method are further corroborated during the complex task of aerial vent pipe inspection, as shown in Fig. \ref{fig:tunnel}, reiterating its potential for various practical applications.

The benchmark trajectory estimation results for scenarios involving failure and recovery, wall inspection, and mixed camera types are provided in Tab. \ref{tab:traj_benchmark}. Examination of the table reveals that the proposed method outperforms single-camera approaches in terms of accuracy across all three scenarios, illustrating the advantage of employing multiple cameras. Additionally, the proposed method surpasses the performance of the ablation of dynamic feature allocation version in terms of accuracy, further proving the efficacy of the strategy.

\section{Conclusion}
\label{sec:conclusion}

In this paper, we propose a robust feature-aware multi-camera-IMU state estimator for asynchronous camera modules. The estimator encompasses parallel front ends, a front end coordinator and a back end optimization. It makes efficient use of input frames by implementing a dynamic feature number allocation and a frame priority coordination strategy. The superior robustness and performance of the estimator is validated through real-flight experiments in various challenging scenarios.  

\newlength{\bibitemsep}\setlength{\bibitemsep}{.0238\baselineskip}
\newlength{\bibparskip}\setlength{\bibparskip}{0pt}
\let\oldthebibliography\thebibliography
\renewcommand\thebibliography[1]{%
  \oldthebibliography{#1}%
  \setlength{\parskip}{\bibitemsep}%
  \setlength{\itemsep}{\bibparskip}%
}

\bibliography{ICRA2024_Luqi} 
\end{document}